\documentclass[10pt,twocolumn,letterpaper]{article}

\usepackage{cvpr}              

\usepackage[dvipsnames]{xcolor}

\usepackage[accsupp]{axessibility} 
\definecolor{cvprblue}{rgb}{0.21,0.49,0.74}
\usepackage[pagebackref,breaklinks,colorlinks,citecolor=cvprblue]{hyperref}

\def\paperID{14973} 
\def\confName{CVPR}
\def\confYear{2024}

\iffalse 
    \newcommand{\holger}[1]{\noindent}
    \newcommand{\ww}[1]{\noindent}
    \newcommand{\yclin}[1]{\noindent}
\else
    \newcommand{\holger}[1]{[{\bf \color{orange} HC: #1}]}
    \newcommand{\yclin}[1]{[{\bf \color{blue} YL: #1}]}
\fi

\newcommand{\Eq}[1]{Eq.~(\ref{eq:#1})}
\newcommand{\eq}[1]{\Eq{#1}}
\newcommand{\fig}[1]{Fig.~\ref{fig:#1}}
\newcommand{\tab}[1]{Tab.~\ref{tab:#1}}

\newcommand{\sect}[1]{Section~\ref{sec:#1}}

\def\argmin{\operatornamewithlimits{\rm arg\,min}}

\usepackage{arydshln}
\usepackage{siunitx}
\usepackage{xcolor}
\usepackage{graphicx}
\usepackage{amsmath}
\usepackage{mathtools}  
\usepackage{amsfonts} 
\usepackage{amssymb}
\usepackage{bbm}
\usepackage{dsfont}
\usepackage{booktabs}
\usepackage{comment}
\usepackage{float} 
\usepackage{dblfloatfix} 
\usepackage{pifont}
\newcommand{\xmark}{\ding{55}}%

\newcommand{\norm}[1]{\left\lVert#1\right\rVert}

\usepackage{xspace}

\begin{document}

\title{ICP-Flow: LiDAR Scene Flow Estimation with ICP}

\author{
Yancong Lin and Holger Caesar \\
Delft University of Technology, The Netherlands\\
}
\maketitle

\begin{abstract}

Scene flow characterizes the 3D motion between two LiDAR scans captured by an autonomous vehicle at nearby timesteps.
Prevalent methods consider scene flow as point-wise unconstrained flow vectors that can be learned by either large-scale training beforehand or time-consuming optimization at inference. 
However, these methods do not take into account that objects in autonomous driving often move rigidly. 
We incorporate this rigid-motion assumption into our design, where the goal is to associate objects over scans and then estimate the locally rigid transformations. 
We propose ICP-Flow, a learning-free flow estimator.
The core of our design is the conventional Iterative Closest Point (ICP) algorithm, which aligns the objects over time and outputs the corresponding rigid transformations. Crucially, to aid ICP, we propose a histogram-based initialization that discovers the most likely translation, thus providing a good starting point for ICP. The complete scene flow is then recovered from the rigid transformations.
We outperform state-of-the-art baselines, including supervised models, on the Waymo dataset and perform competitively on Argoverse-v2 and nuScenes. 
Further, we train a feedforward neural network, supervised by the pseudo labels from our model, and achieve top performance among all models capable of real-time inference. 
We validate the advantage of our model on scene flow estimation with longer temporal gaps, up to 0.4 seconds where other models fail to deliver meaningful results.

\end{abstract}    
\section{Introduction}
\label{sec:intro}

\begin{figure}
    \centering
    \includegraphics[width=0.5\textwidth]{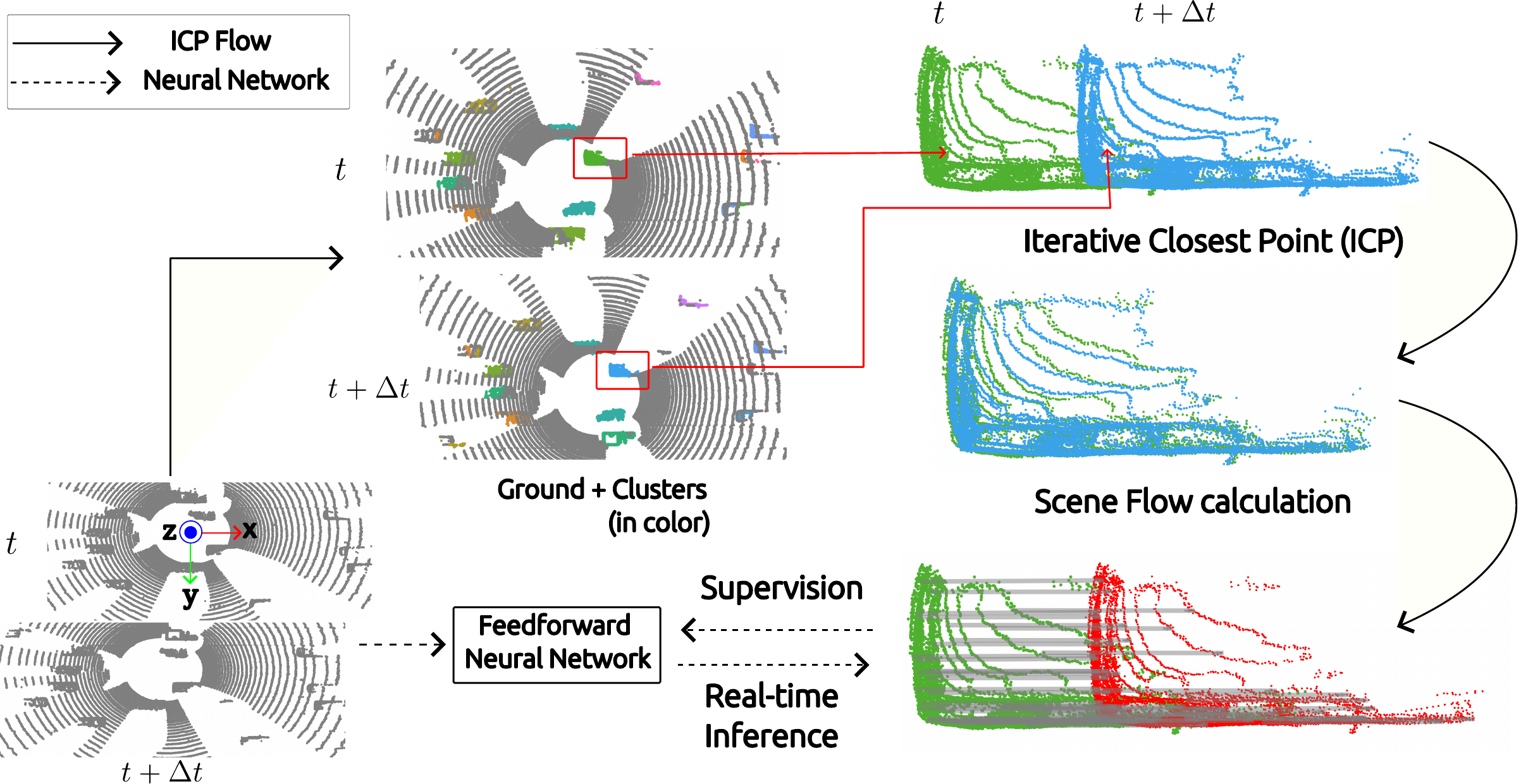}
    \caption{\textbf{ICP for scene flow.}  
    Given two LiDAR scans, we remove the ground, cluster points, and align clusters using ICP, as objects move rigidly. We infer a rigid transformation for each pair of clusters, from which the scene flow can be recovered. 
    Further, we train a feedforward network using the prediction from our model as supervision. The network runs in real-time with only marginal performance loss.
    }
    \label{fig:intro}
\end{figure}

Motion is vital for visual perception, particularly for highly automated vehicles that operate in a dynamically changing 3D world, as motion facilitates the detection of dynamic objects around an autonomous vehicle. A popular task in motion prediction is scene flow estimation, which calculates point-wise motion from two temporarily adjacent LiDAR scans, \textit{i.e.} a 3D vector that describes the displacement of a point~\cite{dewan2016rigid,ushani2017learning,behl2019pointflownet, liu2019flownet3d}. Scene flow lays the foundation for numerous high-level tasks in perception, particularly in scene understanding without relying on large amounts of annotations~\cite{zhai2020flowmot,  najibi2022motion, erccelik20223d, wang20224d}. 
As an example, ~\cite{najibi2022motion} leverages scene flow to segment dynamic objects from a scene and associate them over multiple frames, from which one can create bounding boxes for training object detectors in an unsupervised manner.
~\cite{wang20224d} builds on top of scene flow and takes advantage of motion cues to discover and track objects from a large volume of unlabeled data. These works treat scene flow estimation as a cornerstone and consider motion as a useful prior for temporal perception.
The motion prior not only reduces the dependency on manual annotation but also scales up when ample data is available, particularly in autonomous driving where ample unlabeled data is relatively cheap to acquire.
Thus, it is important to develop a reliable scene flow method for autonomous driving.

There has been a strong demand for unsupervised scene flow, as many works~\cite{zhai2020flowmot, najibi2022motion, erccelik20223d, wang20224d} count on scene flow to extract object information, such as tracked bounding boxes over time, for free, \textit{i.e.}, without human labeling. 
Recent works in~\cite{baur2021slim,mittal2020just} have made great strides towards this direction, by utilizing the cycle consistency of forward and backward flows.
Further, the work in~\cite{li2021neural, li2023fast} proposes test-time optimization that conducts learning from scratch per sample, thus eliminating the need for training data.
Unfortunately, these approaches are only able to predict free-form unconstrained scene flow, due to the lack of multi-body rigidity, \textit{i.e.} a scene is composed of multiple rigidly-moving objects. As a consequence, the flow vectors from the same object, \textit{e.g.}, a moving vehicle, may not agree in terms of their direction or magnitude. Although ~\cite{gojcic2021weakly, huang2022dynamic} have done pilot work on incorporating the motion rigidity into scene flow, they rely on either partial or full annotation for model training. 
Recent work \cite{li2022rigidflow} achieves unsupervised learning without losing motion rigidity. We share the same spirit as \cite{li2022rigidflow} and further eliminate the need for large-scale data and lengthy training processes.

Another concern for scene flow is the inference cost, which is crucial for processing large volumes of data, particularly in autonomous driving.
However, recent works~\cite{li2021neural, chodosh2023re, vidanapathirana2023mbnsf}, although being data-independent, suffer from significant inference latency. 
Processing a single sample can take more than a minute on a modern GPU~\cite{li2021neural, chodosh2023re}, making scene flow a time-consuming and resource-intensive task in real-world deployment.
Other unsupervised work~\cite{li2022rigidflow} is unable to process full LiDAR scan during inference and requires downsampling due to high demand on GPU memory.

We propose ICP-Flow, a learning-free model to overcome the reliance on data and the lack of motion rigidity. ICP-Flow also provides high-quality pseudo labels for training a neural network that runs in \textit{real time} at inference.
Our model, as shown in \fig{intro}, builds on top of the Iterative Closest Point (ICP)~\cite{besl1992method} algorithm and is fully hand-crafted, thus demanding neither human annotation nor training data. Although simple, we are able to achieve competitive performance on common benchmarks, including Waymo~\cite{waymo}, Argoverse-v2~\cite{Argoverse2} and nuScenes~\cite{caesar2020nuscenes}. 
Further, we treat our predictions as pseudo-labels for supervising feedforward neural networks and achieve \textit{real-time inference} with only a marginal performance decrease. 
Additionally, we extend our ICP-Flow to scene flow estimation over a longer temporal horizon of up to 0.4 seconds, where other models fail.

To summarize, our contributions are as follows:
\begin{itemize}
\item We introduce a learning-free LiDAR scene flow estimator that requires neither large datasets nor manual annotation.

\item Our ICP-Flow incorporates the multi-body rigid-motion assumption by design and produces consistent scene flow per object.
ICP-Flow is the top-performing model on Waymo \cite{waymo} and nuScenes~\cite{caesar2020nuscenes}.

\item Our ICP-Flow generates high-quality pseudo labels for supervising a feedforward neural network that performs on-par with the state-of-the-art, but with a considerably lower inference latency.
\end{itemize}

\section{Related work}
\label{sec:related_work}

There have been numerous works that estimate scene flow from RGB or RGBD images~\cite{vogel2013piecewise, vogel20113d, wedel2011stereoscopic, menze2015object}. However, our focus is on scene flow from \textit{point clouds}, particularly in \textit{autonomous driving}. We highlight works within this scope.

\subsection{Scene flow from point clouds}
Early work~\cite{dewan2016rigid} on scene flow formulates the task as an energy minimization problem by assuming geometric constancy and motion smoothness.
\cite{ushani2017learning} converts point clouds into occupancy grids and computes a flow field by tracking the occupancy grids using expectation maximization.
Recent works are mostly data-driven models that estimate scene flow in an end-to-end fashion~\cite{behl2019pointflownet,liu2019flownet3d,puy20flot,huang2022dynamic,gu2019hplflownet,liu2019meteornet,kittenplon2021flowstep3d,wang2021festa,cheng2022bi,wang2020flownet3d++}. However, model training requires massive data labeled by human experts. In contrast, our model is free from explicit learning and costly annotation.

To remedy the need for manual labels, \cite{Baur2021ICCV, tishchenko2020self, mittal2020just} take advantage of the cycle consistency and propose a self-supervised mechanism for model training. \cite{wu2020pointpwc} achieves the same goal by minimizing the Chamfer distance between two point clouds after flow compensation, with smoothness constraint and Laplacian regularizer. Instead of self-supervised learning, \cite{jin2022deformation} develops a synthetic dataset with ground truth annotations to aid learning. 
Another line of research focuses on knowledge distillation from imperfect pseudo labels~\cite{vedder2023zeroflow,li2021self}. \cite{vedder2023zeroflow} supervises model training using predictions from~\cite{li2021neural} and is able to outperform the teacher model~\cite{li2021neural} when a sufficient amount of data is available.
Although manual labels are no longer needed, these models still demand large amounts of data. In contrast, our model requires neither training data nor human labeling.

Recently, runtime optimization has gained popularity because of its independence on data. \cite{li2021neural,li2023fast, chodosh2023re,lang2023scoop} learn scene flow for each sample at test time by iteratively minimizing the Chamfer distances between two point clouds. Our model shares the same spirit and eliminates the need for data. However, our design is hand-crafted, thus free from the time-consuming test time optimization.

\subsection{Motion rigidity in scene flow}
Instead of predicting an unconstrained free-form flow vector per point, \cite{dewan2016rigid} uses the rigid body assumption in scene flow, \textit{i.e.} objects do not deform, and predicts a rigid transformation per object.
Similarly, \cite{gojcic2021weakly, huang2022dynamic, dong2022exploiting, li2022rigidflow} also adopt the rigidity assumption by design. 
\cite{vidanapathirana2023mbnsf} considers rigidity as an additional regularizer and improves upon previous work~\cite{li2021neural}, which only produces an unconstrained flow field.
Inspired by this line of research, we also convert scene flow into rigid transformation estimation from which the complete scene flow can be recovered. 
Notably, previous work \cite{li2022rigidflow} has proposed using ICP to align LiDAR segments where the initial transformation of ICP is estimated from a deep network trained on large-scale datasets.
In contrast, we eliminate the need for computationally expensive training of deep networks on large datasets by using a hand-crafted histogram-based scheme to aid ICP.

\subsection{ICP}
ICP~\cite{besl1992method} is a commonly used technique for registering 3D shapes or point clouds, based on point correspondences. 
There have been numerous works on extracting reliable correspondences, ranging from classic feature engineering~\cite{chen1992object,rusinkiewicz2001efficient,rusu2009fast,park2017colored,yang2020teaser, vizzo2023kiss} to deep feature learning~\cite{wang2019deep,choy2020deep,zhang20233d}. 
However, they are primarily designed to match scene-scale data, \textit{e.g.}, full-size LiDAR scans, rather than individual segments from LiDAR data. 
We opt for a conventional ICP implementation~\cite{besl1992method} to match clustered LiDAR segments. 

\begin{figure*}
    \centering
    \includegraphics[width=0.9\textwidth]{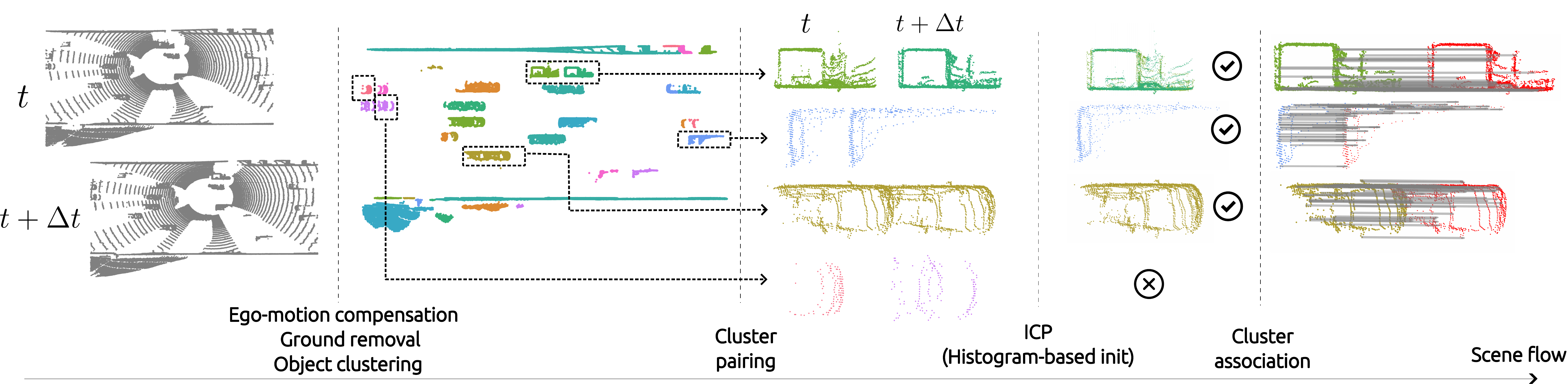}
    \caption{\textbf{Overview of ICP-Flow.} 
    Given two full-size LiDAR scans as input, we first do ego-motion compensation and ground removal on each scan. Subsequently, we fuse the non-ground points from both scans and group them into a set of clusters. We pair clusters by spatial locality and feed them to ICP matching for further verification and transformation estimation. We then filter unreliable matches and associate clusters over time. The scene flow is recovered by using the rigid-motion assumption. Crucially, to aid ICP matching, we develop a histogram-based voting strategy for initialization, by exploring the motion rigidity.
    }
    \label{fig:overview}
\end{figure*}

\section{Method}
\label{sec:method}

\subsection{Problem statement}
Scene flow estimation takes as input a pair of LiDAR scans $\textbf{X}^{t}$ and $\textbf{X}^{t+\Delta t}$, captured by an autonomous vehicle at two adjacent time steps $t$ and $t+\Delta t$, where $\textbf{X} \in {\mathbb{R}^{3\times L}} = 
\{ \textbf{x}_{l} \in \mathbb{R}^{3} \}_{l=1}^{L} $ denotes a point cloud of length $L$.
The goal is to estimate a flow field $\textbf{F}^{t} \in {\mathbb{R}^{3\times L}} = \{ \textbf{f}_{l} \in \mathbb{R}^{3} \}_{l=1}^{L} $ such that $\textbf{X}^{t} + \textbf{F}^{t}  \approx \textbf{X}^{t+\Delta t}$. Notably, the size of $\textbf{X}^{t}$ may differ from $\textbf{X}^{t+\Delta t}$, while $\textbf{X}^{t}$ and $\textbf{F}^{t}$ are always of the same length. 

Additionally, we assume that each LiDAR scan $\textbf{X}$ can be decomposed into \textit{background} $\textbf{X}_{\text{bg}}$ that is static over time and \textit{foreground} $\textbf{X}_{\text{fg}}$ consisting of $K$ rigidly-moving objects that may or may not move at the given time step, denoted as $\textbf{X}_{\text{fg}} = \{ \textbf{C}_{k} \in \mathbb{R}^{3\times L_{k}} \}_{k=1}^{K}$, where $\textbf{C}_{k}$ is the $k$\textit{-th} object, \textit{i.e.}, a cluster of points that represents a particular object.
Our aim is to estimate a rigid transformation $\textbf{T}_{k} \in SE(3)$ for each object $\textbf{C}_{k} $, from which we can recover its scene flow $\textbf{F}_{k} $ by 
\begin{align}
\textbf{F}_{k} = \textbf{T}_{k} \textbf{T}_{ego}  \circ \textbf{C}_{k} - \textbf{C}_{k},
\label{eq:flow_fg}
\end{align}
where $\circ$ indicates applying a rigid transformation to a set of points.
Matrix $\textbf{T}_{ego} \in SE(3)$ is the ego-motion transformation at the corresponding time step.
$\textbf{X}_{\text{bg}}$ is static over time and therefore its transformation $T_{\text{bg}}$ is equivalent to an identity matrix $\textbf{T}_{eye} \in SE(3)$.
Thus, our end goal is to decompose a scene into a set of clusters $\{ \textbf{C}_{k} \}_{k=1}^{K+1}$ representing the objects, and then to estimate their transformations $\{ \textbf{T}_{k} \}_{k=1}^{K+1}$ between two scans $\textbf{X}^{t}$ and $\textbf{X}^{t+\Delta t}$. For simplicity, we consider the \textit{background} as an additional object that does not move over time.

\subsection{Overview of ICP-Flow}
\fig{overview} shows a sketch of ICP-Flow. Given two LiDAR scans, we first conduct ego-motion compensation to align them in the same coordinate system. 
Sequentially, we remove the ground points from each scan separately and fuse the remaining points for subsequent clustering, resulting in a set of clusters (\textit{i.e.} clusters of points) at time $t$ and $t+\Delta t$, respectively. 
We then employ the Iterative Closest Point (ICP)~\cite{besl1992method} algorithm to associate clusters over time. Notably, rather than matching two LiDAR scans, we apply ICP to each pair of clusters over time and estimate a transformation matrix that minimizes the point-wise distance between paired clusters.
Ultimately, we calculate a scene flow per cluster and assign it to the corresponding points in the original LiDAR scan.
Crucially, we highlight that naive ICP does not deliver competitive results because it relies to a great extent on a good initial guess of the transformation.
To overcome this issue, we design a simple yet effective histogram-based initialization for ICP.

Moreover, we train a feedforward neural network to further reduce the inference latency, using the pseudo-ground truth generated by our model.


\subsection{Ego motion Compensation}
Ego motion compensation can significantly reduce the difficulty in scene flow estimation, as the \textit{background} and static objects no longer ``move" after compensation. Further, ego motion is directly available in autonomous driving (from the IMU or other odometry) and in common benchmarks~\cite{waymo, caesar2020nuscenes, Argoverse2}.
Thus, we take advantage of the given ego motion in our design. 
For benchmarks where ego motion is not available, we adopt KISS-ICP~\cite{vizzo2023kiss} to estimate the relative transformation between a pair of scans.

\subsection{Ground removal and point clustering}
We use Patchwork++~\cite{lee2022patchwork++} to remove the ground from each LiDAR scan and feed the remaining points into HDBSCAN~\cite{campello2013density} for clustering. 
Instead of clustering each scan individually, we first fuse the non-ground points from both scans and then conduct HDBSCAN clustering. 
Afterward, we separate fused points by time and obtain a set of clusters $\{ \textbf{C}_{m}^{t} \}_{m=1}^{M}$ and $\{ \textbf{C}_{n}^{t+\Delta t} \}_{n=1}^{N}$, each of which has a timestamp and a cluster index as denoted by the superscript and subscript.

\subsection{Cluster pairing}
\label{sec:pairing}
For each cluster at time $t$, we search for several candidate clusters at $t+\Delta t$ that are likely to match the given cluster. This is to reduce the search space and does \textit{not} enforce one-to-one correspondence. We refer to this step as cluster pairing, after which all paired clusters are fed to ICP matching (\sect{icp_matching}) for further verification (\sect{cluster_association}). If successful, we save the transformation $\textbf{T}_{m}^{t}$ that best aligns each pair of clusters for scene flow calculation.
Simply speaking, we pair each cluster at time $t$ with its neighboring clusters at time $t+\Delta t$ that lie in a predefined area of $\tau_{x} \times \tau_{y}$ where $\tau_{x}$ and $\tau_{y}$ (in meters) are the maximal translations possible within $\Delta t$ along the $x$ and $y$ dimension, respectively. 
Subsequently, we feed all pairs to ICP matching and cluster association if successful.
This procedure can be further simplified by exploiting the clustering indices from HDBSCAN. We provide details in the supplementary material.

\subsection{ICP matching}
\label{sec:icp_matching}
Given a pair of clusters $\textbf{C}_{m}^{t}$ and $\textbf{C}_{n}^{t+ \Delta t}$ consisting of $L_{m}$ and $L_{n}$ points respectively, we leverage ICP~\cite{besl1992method} to estimate a transformation matrix $\textbf{T}_{m}^{t}$ and calculate quantitative metrics to measure the alignment. However, ICP requires a reasonably good initialization; otherwise, it produces a suboptimal estimation. We refer the readers to the ablation study in the supplementary material for an in-depth analysis. 

\noindent \textbf{Histogram-based initialization.} We take advantage of the fact that objects rarely change directions sharply within a short temporal window $\Delta t$, where $\Delta t \leq 0.1$ seconds, leaving translation the major variable to infer during initialization. 
Moreover, we explicitly incorporate the rigid motion assumption, indicating that points from the same object share approximately the same translation within a time gap of $\Delta t$.
Taking this inspiration, we compute a histogram for all translation vectors between a pair of clusters and then select the dominant translation shared by the majority.
We construct the histogram $\textbf{H}$ by discretizing the maximal translation $\tau_{x}$ within $\Delta t$ along the $x$ dimension into equally spaced bins of 0.1 meters. The same also applies to the $y$ and $z$ dimensions. This results in a histogram $\textbf{H}$ of size $L_{x} \times L_{y} \times L_{z}$.
To conduct voting, we calculate the \textit{point-wise} translation vectors between $\textbf{C}_{m}^{t}$ and $\textbf{C}_{n}^{t+ \Delta t}$\footnote{This is equivalent to a broadcasted matrix subtraction.}, namely $\{ \textbf{x}_{i}^{t} - \textbf{x}_{j}^{t+\Delta t} \; | \; \textbf{x}_{i}^{t} \in \textbf{C}_{m}^{t}, \; \textbf{x}_{j}^{t+\Delta t}\in \textbf{C}_{n}^{t+\Delta t} \}$ where $i = 1, \ldots, L_{m}$, and $j = 1, \ldots, L_{n}$.
Subsequently, we discretize each translation vector and cast a vote to its bin in $\textbf{H}$. 
After voting, we localize the bin with the most votes and initialize $\textbf{T}_{m}^{t}$ with the associated translation. 

\noindent \textbf{ICP matching.} We further refine the initial transformation using ICP~\cite{besl1992method}.
We opt for a conventional ICP implementation rather than recent works that are primarily designed for large-scale point clouds and require advanced GPUs for optimization. 
We adopt the Pytorch3D implementation~\cite{ravi2020pytorch3d}. 

To evaluate the quality of ICP matching, we adopt two metrics: average distance $d$ between transformed point correspondences and ratio of inliers $r$, defined by $d = \frac{1}{L_{m}} \sum_{i=1}^{L_{m}} \norm{ d_{i}}$ and $r = \frac{ \sum_{i=1}^{L_{m}} \mathds{1}(d_{i})} {L_{m}+L_{n}-\sum_{i=1}^{L_{m}} \mathds{1}(d_{i})}$, respectively.
Here $ d_{i}$ indicates the distance between a point $\textbf{x}_{i}^{t} \in \textbf{C}_{m}^{t}$ and its nearest neighbor $\textbf{x}_{j}^{t+\Delta t}\in \textbf{C}_{n}^{t+\Delta t}$ after transformation. $\mathds{1}(d_{i})$ is an indicator function, defined by \eq{nn_inlier}, that categorizes a pair of correspondences as an inlier if the distance does not exceed $\tau_{inlier}$:

\begin{align}
    \vspace{-6mm}
    \mathds{1}(d_{i}) =     
    \begin{cases}
      1 &  \text{$ \norm{d_{i}} \leq \tau_{inlier}$  }\\
      0 & \text{otherwise}
    \end{cases}   
\label{eq:nn_inlier}
\end{align}

\subsection{Cluster association}
\label{sec:cluster_association}
After ICP matching, we construct a distance matrix $\textbf{M}_{d}$ of size $L_{m} \times L_{n}$, where $\textbf{M}_{d}(m, n)$ indicates the distance $d$ between paired clusters $\textbf{C}_{m}^{t}$ and $\textbf{C}_{n}^{t+\Delta t}$. Similarly, we also build an inlier-ratio matrix $\textbf{M}_{r}$ where $\textbf{M}_{r}(m, n)$ indicates the inlier ratio $r$ between paired clusters. For unpaired clusters, we assign $\infty$ in $\textbf{M}_{d}$ and $0$ in $\textbf{M}_{r}$. 

Sequentially, for each query cluster at time $t$, we seek a paired cluster at time $t+\Delta t$ that has the smallest distance $d$. However, it might be possible that a pair of clusters differ substantially in size, such that the cluster with fewer points always matches well with the other. To prevent this, we reject a pair of clusters once $r < \tau_{r}$ where $\tau_{r}$ is a predefined threshold. 
Afterwards, we search for the best match for the query cluster.
This is equivalent to an $\argmin$ over columns for each row in $\textbf{M}_{d}$. Similarly, we reject paired clusters once $d>\tau_{d}$, where $\tau_{d}$ is also a threshold.
For clusters at time $t$ that have no match at time $t+\Delta t$, we simply assign an identity transformation. Finally, we recover the scene flow using \eq{flow_fg}.

Alternatively, we can apply Hungarian matching~\cite{crouse2016implementing} to associate the clusters from two scans, by enforcing one-to-one correspondence. In general, we find that $\argmin$ matching works well on common benchmarks.

\subsection{ICP-Flow pseudo labels as supervision}
We also train a feedforward neural network for real-time inference. We supervise model training with the pseudo labels from ICP-Flow.
We adopt the same setup as ZeroFlow~\cite{vedder2023zeroflow}, including both the model architecture and the loss function.

\subsection{Implementation details}
We adopt the default parameters in Patchwork++~\cite{lee2022patchwork++}
during ground removal. 
We use the default parameters in HDBSCAN~\cite{mcinnes2017hdbscan, mcinnes2017accelerated}, except that $min\_cluster\_size$ is set to 20, below which ICP matching becomes substantially harder.
We take maximally 200 clusters after HDBSCAN sorted by the number of points for cluster pairing. 
For the other clusters, we simply set their transformations to be an identity matrix.
Assuming $\Delta t = 0.1 $ seconds, we set the maximal translation $\tau_{x}$ and $\tau_{y}$ to be 3.33 meters, which is equivalent to the distance that an agent travels at 120 km/h. Scalar $\tau_{z}$ is set to be 0.1 meters, as objects barely move up/downward. 
We set the inlier threshold $\tau_{inlier}$ during ICP~\cite{zhou2018open3d, ravi2020pytorch3d} to be 0.1 meters. We set the rejection threshold for cluster association to be $\tau_{d}=0.2$ meters or $\tau_{r}=0.2$.  

Regarding neural network training, we use the Adam optimizer~\cite{kingma2014adam} with an initial learning rate of $2e^{-4}$, which is multiplied by $0.1$ after 25 epochs. We train the model for 50 epochs on 4 Nvidia V100 GPUs and an Intel(R) Xeon(R) W-2245 CPU @ 3.90GHz. The entire training process takes approximately 5 days on the Waymo scene flow dataset~\cite{huang2022dynamic, waymo}. We conduct model inference on the same device.
Our code is available at~\url{https://github.com/yanconglin/ICP-Flow}.

\section{Experiments}
\label{sec:exp}

\begin{table*}
    \centering
    \resizebox{1.0\linewidth}{!}{
    \begin{tabular}{l cc ccc ccc ccc}
        \toprule
       Metrics & Label & Time & \multicolumn{3}{c}{Dynamic Foreground} & \multicolumn{3}{c}{Static Foreground} & \multicolumn{3}{c}{Static Background} \\
        \cmidrule(lr){4-6} \cmidrule(lr){7-9}\cmidrule(lr){10-12}
          & & & EPE ($m$) $\downarrow$ & Acc-S ($\%$) $\uparrow$ & Acc-R ($\%$) $\uparrow$ & EPE ($m$) $\downarrow$ & Acc-S ($\%$) $\uparrow$ & Acc-R ($\%$) $\uparrow$ & EPE ($m$)$\downarrow$ & Acc-S ($\%$) $\uparrow$ & Acc-R ($\%$) $\uparrow$ \\ \cmidrule(lr){1-12}
         \cmidrule(lr){1-12}
        NSFP~\cite{li2021neural} & - & 90s & 0.0966 & 61.12 & 79.64  & 0.0162 & 92.26 &  98.74 & 0.0538 & 89.56 & 93.32    \\
        \small{Chodosh \textit{et al.}}~\cite{chodosh2023re} &-  & 93s &  0.1081 & 59.15 & 78.30  & 0.0156 & 95.54 & 98.80 & 0.0448 & 91.35 & 95.46  \\ 
        FastNSF~\cite{li2023fast} & - 
 & 0.5s& 0.2983 & 32.06 & 46.84 & 0.0146 & 97.69 & 99.30 & 0.0402 & 94.39 & 96.73   \\ 
        
        ZeroFlow~\cite{vedder2023zeroflow} & - & \textbf{21ms}  & 0.2229 & 15.20 & 36.56 & 0.0123 & 96.30 & 98.46 & 0.0198 & 96.75 & 97.84  \\
        RigidFlow~\cite{li2022rigidflow} & - & 0.6s  & 0.2575 & 15.99 & 38.29 & 0.1299 & 30.11 & 58.37 & 0.2517 & 18.08 & 40.15 \\
      \cmidrule(lr){1-12}
      
        FastFlow~\cite{jund2021scalable} & \checkmark & \textbf{21ms} & 0.1950 & 25.44 & 49.55  & 0.0170 & 89.17 & 96.19 & \textbf{0.0031} & 99.15 & 99.50  \\
        PCA~\cite{huang2022dynamic} & \checkmark  & 35ms & 0.1083 & 61.93 & 83.41  & 0.0156 & 98.35 & 97.78 & 0.0199 & 98.12 & 99.56   \\ \cmidrule(lr){1-12}

        Ours &  - & 3.3s & \textbf{0.0799} & \textbf{76.10} &  \textbf{88.53} & 0.0165 & 96.46 & \textbf{99.68}  & 0.0270 & 96.83 & 98.51  \\ 
        Ours+FNN & -& \textbf{21ms} & 0.1254 & 41.22 & 67.31 & \textbf{0.0044} & \textbf{98.91} & 99.45  & 0.0058 & \textbf{99.58} & \textbf{99.62} \\ 
        \bottomrule
    \end{tabular}}
    \caption{
        \textbf{Comparison on Waymo dataset~\cite{waymo,huang2022dynamic}}. 
        This dataset contains paired LiDAR scans from successive time steps, \textit{after ego motion compensation}. 
        We evaluate all methods using EPE, Acc-S, and Acc-R, on dynamic foreground, static foreground, and static background separately. 
        Overall, our model and its derivatives perform the best over multiple metrics. Notably, we are also able to outperform supervised baselines, particularly for the dynamic foreground.
        Among all methods, ZeroFlow, FastFlow, and Ours+FNN have identical model designs, thus having the same inference speed. 
        }
    \label{tab:waymo}
\end{table*}

\begin{table*}
    \centering

    \resizebox{1.0\linewidth}{!}{
    \begin{tabular}{l cc ccc ccc ccc}
        \toprule
       Metrics & Label & Time & \multicolumn{3}{c}{Dynamic Foreground} & \multicolumn{3}{c}{Static Foreground} & \multicolumn{3}{c}{Static Background} \\
        \cmidrule(lr){4-6} \cmidrule(lr){7-9}\cmidrule(lr){10-12}
          & &  & EPE ($m$) $\downarrow$ & Acc-S ($\%$) $\uparrow$ & Acc-R ($\%$) $\uparrow$ & EPE ($m$) $\downarrow$ & Acc-S ($\%$) $\uparrow$ & Acc-R ($\%$) $\uparrow$ & EPE ($m$)$\downarrow$ & Acc-S ($\%$) $\uparrow$ & Acc-R ($\%$) $\uparrow$ \\ \cmidrule(lr){1-12}

        NSFP~\cite{li2021neural}  & - & 26s & 0.1328 & 48.84  & 70.97  & 0.0343 & 87.65 & 96.37 & 0.0371 & 89.19 &  95.92  \\
        \small{Chodosh \textit{et al.}}~\cite{chodosh2023re}  & - & 35s &  \textbf{0.1311} & \textbf{49.40} & \textbf{71.78} & 0.0261 & 89.52 & 97.15 & 0.0213 & 91.37 & 96.51\\ 
        FastNSF~\cite{li2023fast}& - & 0.5s & 0.3684 & 23.60 & 39.24 &  0.0227 & 94.65 & 99.02 & 0.0268 & 93.13 & 98.19   \\ 

        ZeroFlow~\cite{vedder2023zeroflow} & - & \textbf{29ms}  & 0.2244 & 15.82 & 40.04 & 0.0205 & 94.67 & 97.80  & 0.0125 & 97.62 & 98.88   \\
        \cmidrule(lr){1-12}
        FastFlow~\cite{jund2021scalable} & \checkmark & \textbf{29ms}  & 0.1836 & 22.52 & 50.67  &  0.0198 & 92.49 & 96.86 & 0.0064 & 98.38 & 99.09  \\

        \cmidrule(lr){1-12}
        Ours      & -  & 3.5s   & 0.1653 & 48.61 &  70.70  & 0.0391 & 79.31 & 95.32 & 0.0320 & 86.68 & 95.68 \\ 
        Ours+FNN     & -  &  \textbf{29ms} & 0.1697 & 36.01 & 62.88 &  \textbf{0.0189} & \textbf{94.76} & \textbf{99.22} & \textbf{0.0035} & \textbf{99.37} & \textbf{99.61}   \\
        \bottomrule

    \end{tabular}}
    \caption{
        \textbf{
        Comparison on Argoverse-v2 dataset~\cite{Argoverse2, chodosh2023re, vedder2023zeroflow}.} 
        The dataset contains pairs of successive LiDAR scans \textit{after ego motion compensation}.
        NSFP~\cite{li2021neural} and Chodosh \textit{et al.} are the state-of-the-art for dynamic foreground. However, they require significantly longer time than others for optimization, up to half a minute. In contrast, Ours+FNN, a feedforward neural network supervised by Ours, is capable of real-time inference without decreasing the performance. Although being less superior on dynamic foreground, Ours+FNN achieves top results on static foreground and background.
    }
    \label{tab:argo}
\end{table*}


\subsection{Datasets}
We conduct experiments on the Waymo~\cite{waymo}, nuScenes~\cite{caesar2020nuscenes} and Argoverse-v2~\cite{Argoverse2} datasets, which are the largest datasets for scene flow in autonomous driving. We take full-size LiDAR scans as input without any downsampling.

\textbf{Waymo.}
We use the \emph{modified} Waymo dataset released by~\cite{huang2022dynamic}, where the ground truth is calculated from annotated 3D bounding boxes. There are 11,440/4,013/4,032 samples for training/validation/test, where each sample consists of 5 consecutive scans spanning 0.4 seconds, as the LiDAR frequency is 10Hz. 
The average number of points per scan is 177,000~\cite{chodosh2023re}. 
We follow~\cite{huang2022dynamic} to remove the ground points, by applying a threshold along the $z$ axis. 

\textbf{nuScenes.}
Similar to Waymo, we also use the \emph{modified} nuScenes dataset from~\cite{huang2022dynamic}. 
There are 10,921/2,973/2,973 samples for training/validation/test, where each sample contains a sequence of 11 consecutive scans captured at 20Hz. Notably, nuScenes is sparser than Waymo due to the sensor difference (32 beams \textit{vs.} 64 beams). The average number of points per scan is 25,000~\cite {chodosh2023re}. We also remove the ground points by thresholding along the $z$ axis~\cite{huang2022dynamic, baur2021slim}. 

We are also aware of the existence of other subsets for Waymo and nuScenes, such as the ones used in~\cite{jund2021scalable, baur2021slim, li2021neural, li2023fast}. We choose the subset from~\cite{huang2022dynamic} because it has $(1)$ abundant samples for test; and $(2)$ paired scans over a longer temporal horizon.

\textbf{Argoverse-v2.}
We adopt the recent Argoverse-v2~\cite{Argoverse2}, captured by two roof-mounted 32-beam LiDARs. This dataset only contains paired LiDAR scans at two successive time steps with an interval of 0.1 seconds. We follow the exact preprocessing procedure as in~\cite{vedder2023zeroflow} and conduct evaluation on the official validation split. The average number of points per scan is 83,000~\cite{chodosh2023re}. The ground points are removed according to a rasterized heightmap.

\subsection{Evaluation}
We adopt three metrics for evaluation~\cite{huang2022dynamic, chodosh2023re}, including 
$(1)$ 3D end-point-error (EPE, in \textit{meters}) which measures the average $L_{2}$ error of all flow vectors; 
$(2)$ strict accuracy (Acc-S, in $\%$), equivalent to the fraction of points with EPE $\le0.05m$ or relative EPE error (to ground truth norm) $\le 0.05$;
$(3)$ relaxed accuracy (Acc-R, in $\%$), similar to Acc-S but with a threshold of $0.1m$ and $0.1$. We report these three metrics on \textit{static foreground}, \textit{static background}, and \textit{dynamic foreground}\footnote{A point is considered as \textit{dynamic} if its ground truth velocity is above $0.5 m/s$.}~\cite{chodosh2023re}. This provides a more comprehensive evaluation than an overall metric averaged over all points, as \textit{static background} points are the majority in a scene.
The evaluation is limited to points within a $64m \times 64 m$ area surrounding the ego-car on Waymo and nuScenes~\cite{huang2022dynamic}. On Argoverse-v2, the evaluation is extended to a $102.4m \times 102.4 m$ area~\cite{vedder2023zeroflow, Argoverse2}.

\subsection{Baselines}
We compare our mode against 5 recent baselines, including RigidFlow~\cite{li2022rigidflow}, NSFP~\cite{li2021neural}, FastNSF~\cite{li2023fast}, FastFlow~\cite{jund2021scalable}, ZeroFlow~\cite{vedder2023zeroflow}, PCA~\cite{huang2022dynamic}. 
RigidFlow~\cite{li2022rigidflow} shares the same strategy as ours except that the initial transformation for ICP is estimated by a pre-trained neural network. We use the released checkpoint by the authors (trained on KITTI$_{r}$~\cite{li2021self}) and report its results. 
NSFP~\cite{li2021neural} and FastNSF~\cite{li2023fast} come from a family of work that employs test-time optimization, with FastNSF being substantially faster, as indicated by its name. Both methods require no training data or manual annotation. ZeroFlow~\cite{vedder2023zeroflow} and FastFlow~\cite{jund2021scalable} are both data-driven methods. The major difference is that FastFlow is supervised by ground truth labels, while ZeroFlow learns from pseudo labels generated by NSFP~\cite{li2021neural}. PCA~\cite{huang2022dynamic} is a fully-supervised data-driven approach that incorporates into its design the multi-body rigidity. 
For ZeroFlow, we directly use the pre-trained checkpoints released by the authors. We choose ZeroFlow-1X as it does not require training on external data~\cite{vedder2023zeroflow}.
For FastFlow, we directly use the checkpoints from~\cite{vedder2023zeroflow}, which has done extensive comparisons between FastFlow and ZeroFlow.
For PCA~\cite{huang2022dynamic}, we take the official checkpoints released by the authors.
For NSFP and FastNSF, we take the official implementation and adapt it to corresponding datasets. We keep the default parameters, except that weight decay is disabled~\cite{chodosh2023re}.
For Chodosh \textit{et al.}~\cite{chodosh2023re}, we use a third-party implementation\footnote{\url{https://github.com/kylevedder/zeroflow/blob/master/models/chodosh.py}}, as no official code is available. 
We test baseline models on corresponding datasets \textit{by ourselves} due to the lack of certain metrics.


\subsection{Comparison to state-of-the-art}

\paragraph{Waymo. } \tab{waymo} shows the result on the Waymo Open dataset~\cite{waymo, huang2022dynamic}. We compare the EPE, Acc-S, and Acc-R metrics on dynamic foreground, static foreground, and static background separately. 
Our method outperforms not only the unsupervised competitors but also supervised models trained with massive data and annotation, such as PCA~\cite{huang2022dynamic} and FastFlow~\cite{jund2021scalable}, especially on dynamic foreground, \textit{i.e.}, annotated objects that move faster than 0.5m/s. 
Our advantage over the best-performing baseline NSFP~\cite{li2021neural} is approximately 1.5cm per point in terms of EPE.
Notably, our method not only excels in EPE but also improves Acc-S substantially by more than 10$\%$ and Acc-R by 5$\%$. 
Regarding static objects, most models produce reasonably good results and the performance gap among baselines is marginal.
Ours+FNN is a feedforward neural network that shares the same architecture as ZeroFlow~\cite{vedder2023zeroflow}. The only difference is the source of supervision. Ours+FNN is supervised by the pseudo labels from \textit{Ours}, while ZeroFlow uses the NSFP pseudo-labels. Although less competitive than Ours, Ours+FNN still outperforms ZeroFlow by a margin of 10 cm, which shows the value of pseudo-labels generated from our method.
NSFP and Chodosh \textit{et al.}~\cite{chodosh2023re} are also strong competitors in terms of performance, but they are dramatically slow during inference.
A single inference takes more than 1 minute, preventing them from real-world deployment. In contrast, Ours only takes around 3 seconds, thus being approximately $30\times$ faster. Ours+FNN further reduces the runtime by more than $\times 1000$, without sacrificing much performance.
It is worth mentioning that ZeroFlow training requires calculating NSFP pseudo labels beforehand. In this case, NSFP takes several months of GPU compute~\cite{vedder2023zeroflow}, while Ours reduces the effort to several days. 
We also compare Ours to RigidFlow~\cite{li2022rigidflow} as both models follow the ``clustering + ICP" design. The main difference is that RigidFlow requires a deep network for initial pose estimation, while Ours adopts histogram-based initialization. In \tab{waymo}, Ours achieves $\times3$ better result on dynamic foreground and $\times 10$ on static part than RigidFlow in terms of EPE, indicating the usefulness of the proposed initialization. We provide an additional comparison on KITTI$_{o}$\cite{li2021self} in the supplementary material to further validate the advantage of our design.


\paragraph{Argoverse-v2.} We also make comparisons on the recent Argoverse-v2 datasets~\cite{Argoverse2, chodosh2023re}. Chodosh \textit{et al.}~\cite{chodosh2023re} and NSFP~\cite{li2021neural} are the leading methods and outperform others on the dynamic foreground by approximately 3 cm in EPE.
However, as indicated by the running time, they are remarkably slower than others (up to $\times$1000), due to the time-consuming runtime optimization. FastNSF alleviates this issue but suffers from observable performance drops. In spite of the inferiority on dynamic foreground, Ours+FNN, a feedforward neural network supervised by Ours, excels in static foreground and background. More importantly, it enables real-time inference, which is crucial for processing large volumes of data. When compared to ZeroFlow - another unsupervised model that shares the same architecture, Ours+FNN is able to outperform significantly on dynamic foreground (6cm in EPE and 10$\%$ in Acc-S and Acc-R). 
Overall, we are able to achieve the best result among all models that run in real time.


\begin{table*}[t]
    \centering
    \resizebox{1.0\linewidth}{!}{
    \begin{tabular}{lc ccc ccc ccc}
        \toprule
        Metrics  & Label  & \multicolumn{3}{c}{Dynamic Foreground} & \multicolumn{3}{c}{Static Foreground} & \multicolumn{3}{c}{Static Background} \\
         \cmidrule(lr){3-5} \cmidrule(lr){6-8} \cmidrule(lr){9-11}
          &  & EPE ($m$) $\downarrow$ & Acc-S ($\%$) $\uparrow$ & Acc-R ($\%$) $\uparrow$ & EPE ($m$) $\downarrow$ & Acc-S ($\%$) $\uparrow$ & Acc-R ($\%$) $\uparrow$ & EPE ($m$)$\downarrow$ & Acc-S ($\%$) $\uparrow$ & Acc-R ($\%$) $\uparrow$   \\ \cmidrule(lr){1-11}
    
        NSFP~\cite{li2021neural}   &  -   & 0.1527  & 38.82 & 59.46  & 0.0406 & 80.69 & 91.84 & 0.0812 & 66.25 & 79.92   \\

        \small{Chodosh \textit{et al.}}~\cite{chodosh2023re} & - & 0.1571 & 33.27 & 57.17 & 0.0404 & 75.82 & 90.49 & 0.0776 & 63.61 & 80.17   \\ 
                


        FastNSF~\cite{li2023fast}  & - & 0.1591 & 42.63 & 61.43 & 0.0418 & 80.09 & 91.84 & 0.0902 & 58.81 & 77.28  \\ 
        \cmidrule(lr){1-11}
 
        PCA~\cite{huang2022dynamic} & \checkmark & \textbf{0.1340} & 31.89 & 64.91  & 0.0356 & 80.34 & 95.35 & 0.0514 & 64.16 & 88.95  \\ \cmidrule(lr){1-11}
 
       Ours   & -    & 0.1445 & \textbf{49.59} &  \textbf{66.02}  & 0.0298 & 87.26 & 96.22  & 0.0403 & 80.60 & 91.75  \\ 
        Ours+FNN & -    &  0.1850& 21.52 & 45.35 &  \textbf{0.0150} & \textbf{96.05} & \textbf{98.76} & \textbf{0.0090} & \textbf{98.10} & \textbf{98.81}\\
        \bottomrule
    \end{tabular}}
    \caption{
        \textbf{Comparison on nuScenes dataset~\cite{caesar2020nuscenes}. } 
        nuScenes dataset contains paired scans captured by a 32-beam LiDAR sensor. Ours outperforms the unsupervised baselines on all metrics, while getting close to the fully supervised baseline PCA~\cite{huang2022dynamic}.   
    }
   \label{tab:nuscenes}
\end{table*}

\paragraph{nuScenes. } Additionally, we show a comparison on nuScenes~\cite{caesar2020nuscenes}, composed of paired scans every 0.05 seconds, \textit{after ego-motion compensation}. Differing from Waymo and Argoverse-v2, the data is captured by a single 32-beam LiDAR, thus being much sparser. Generally, our model achieves top results in all three categories, compared to other unsupervised baselines, as shown in~\tab{nuscenes}. Notably, Ours is only marginally worse than the supervised baseline PCA~\cite{huang2022dynamic} in terms of EPE for the dynamic foreground.


\begin{table*}
    \centering

    \resizebox{1.0\linewidth}{!}{
    \begin{tabular}{l c ccc ccc ccc}
        \toprule
        Metrics  & Label  & \multicolumn{3}{c}{Dynamic Foreground} & \multicolumn{3}{c}{Static Foreground} & \multicolumn{3}{c}{Static Background} \\
         \cmidrule(lr){3-5} \cmidrule(lr){6-8} \cmidrule(lr){9-11}
          &  & EPE ($m$) $\downarrow$ & Acc-S ($\%$) $\uparrow$ & Acc-R ($\%$) $\uparrow$ & EPE ($m$) $\downarrow$ & Acc-S ($\%$) $\uparrow$ & Acc-R ($\%$) $\uparrow$ & EPE ($m$)$\downarrow$ & Acc-S ($\%$) $\uparrow$ & Acc-R ($\%$) $\uparrow$   \\ \cmidrule(lr){1-11}
        
        NSFP~\cite{li2021neural}   & -  & 0.8533 & 20.22  & 36.37  & 0.0257 & 95.12 & 98.25 & 0.0575 & 93.67 &  96.88   \\
        
        FastNSF~\cite{li2023fast} & - & 4.1415 & 24.72 & 39.15 & 0.0360 & 96.10 & 98.33 &  0.6372 & 93.11 & 95.77   \\ 

        ZeroFlow~\cite{vedder2023zeroflow} & - & 0.7097 & 10.30 & 25.04 & 0.0231 & 94.51 & 97.14 & 0.0340 & 95.71 & 97.17 \\

        FastFlow~\cite{jund2021scalable} & - & 0.6968 & 16.46 & 32.83 & 0.0241 & 88.68 & 95.43 & \textbf{0.0049} & 99.07 & 99.47    \\
        \cmidrule(lr){1-11}
        PCA~\cite{huang2022dynamic} & \checkmark & \textbf{0.1970} & 53.31 & 77.49  & 0.0216 & 97.16 & \textbf{99.44} & 0.0289 & 97.16 & 99.44 \\
       \cmidrule(lr){1-11}
        
        Ours  & - & 0.2209 & \textbf{67.59} & \textbf{84.66}   & 0.0272 & 96.08 & 99.16 & 0.0711 & 96.49 & 97.96\\ 
        Ours+FNN  & - & 0.5636 & 30.76 & 51.73   & \textbf{0.0105} & \textbf{97.44} & 98.54 & 0.0085 & \textbf{99.43} & \textbf{99.51}\\ 
        \cmidrule(lr){1-11}
        Ours+Tracker     & -   &  0.1799 & 58.98 & 80.98  & 0.0341 & 88.53 & 97.73 & 0.0722 & 93.74 & 97.46   \\ 
        \bottomrule

    \end{tabular}}
    \caption{
    \textbf{Scene flow on Waymo dataset~\cite{waymo}, over a longer temporal horizon (5 consecutive frames, up to 0.4 seconds).} 
    Given a clip of 5 consecutive scans, we compute the flow between the first frame and the other frames, leading to 4 pairs per clip. The result is averaged over all points. 
    Most methods fail to generalize to a longer temporal duration, while \textit{Ours} still produces reasonably good results, compared to PCA~\cite{huang2022dynamic} which is specifically designed for this task. Additionally, we also include Ours+Tracker, an extension of our design that utilizes intermediate scans and tracks clusters iteratively over time. It offers better results than the fully supervised PCA. It is worth mentioning that Ours+Tracker takes as input intermediate scans while others do not.
    }
    \label{tab:waymo5}
\end{table*}

\subsection{Scene flow over a longer temporal horizon}
\label{sec:exp_tracker}
So far we compared different methods on scene flow from two successive frames.
We also test the capability of various methods on samples with a longer time difference. This is particularly useful when processing temporally downsampled data~\cite{rempe2020caspr}. Thus we conduct experiments on scene flow estimation from clips of LiDAR scans, each of which contains 5 consecutive scans from Waymo, following~\cite{huang2022dynamic}. 
We calculate scene flow between the first frame and the remaining frames, thus resulting in 4 pairs of LiDAR scans whose time difference gradually increases from 0.1 to 0.4 seconds.

\begin{figure}
    \centering
    \includegraphics[width=0.45\textwidth]{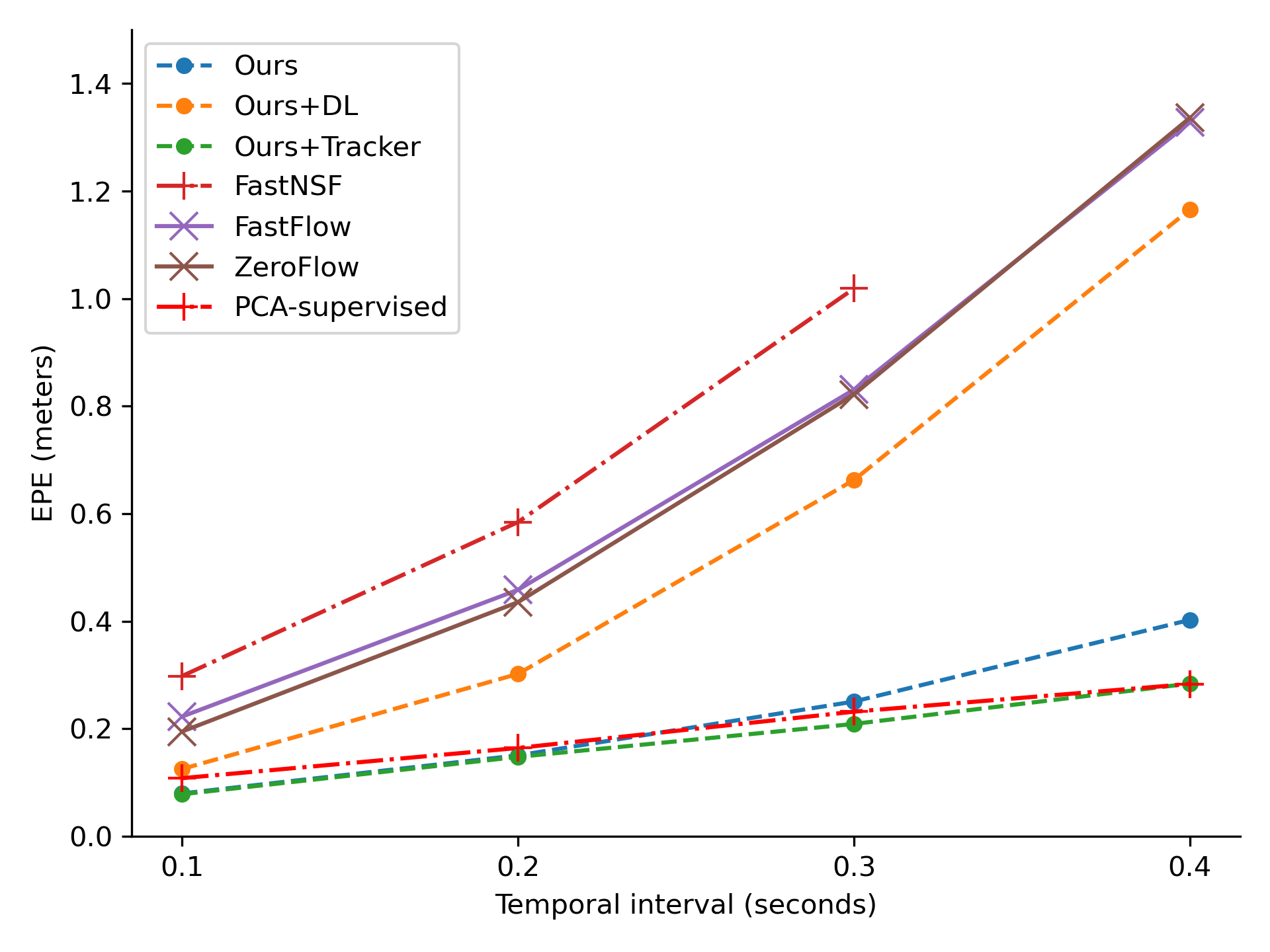}
    \vspace{-2mm}
    \caption{
    \textbf{Scene flow errors with increasing time gap.} 
    We show the EPE values for dynamic foreground with respect to the time duration. As the time gap increases, Ours degrades gracefully and the gap to PCA~\cite{huang2022dynamic}, a supervised model designated for this task, is marginal till 0.3 seconds. In contrast, other methods fail to generalize to a longer duration. Ours+Tracker, an extension of Ours that does tracking over time, is able to achieve comparable results without relying on learning from costly annotation.
    }
    \vspace{-2mm}
    \label{fig:track_waymo}
\end{figure}

We plot the EPE errors over time for the dynamic foreground in \fig{track_waymo}.
With the increase of time, the performance of Ours decreases gracefully. The difference between Ours and PCA~\cite{huang2022dynamic}, a fully supervised data-driven approach on this task, is insignificant within a temporal window of 0.3 seconds. 
In comparison, FastFlow~\cite{jund2021scalable}, ZeroFlow~\cite{vedder2023zeroflow} and Ours+FNN fail to produce a reliable estimation, as the error becomes substantially large, since these methods are not trained for this scenario.  FastNSF~\cite{li2023fast} does not produce reasonable predictions at $t=0.4s$ and is thus absent from comparison. We exclude NSFP~\cite{li2021neural} here because its performance is highly unstable. We are unable to find a set of hyperparameters that work at all time steps. To conclude, our design not only works competitively for scene flow from successive scans but also generalizes to further-away scans within a temporal window of 0.4 seconds. Quantitative comparisons are available in \tab{waymo5}.

We also extend Ours to a tracker, namely Ours+Tracker, which associates clusters over time, \textit{i.e.}, over the entire clip. We provide its details in the supplementary material. In \tab{waymo5}, \textit{Our+Tracker} is able to further improve the result over dynamic foreground by approximately 2cm in EPE as we no longer lose track over time. However, it deteriorates in Acc-S/R, since errors accumulate over time.

\subsection{Limitations}
Our design is a feature-engineering solution that only exploits geometric information during scene flow. However, it can fail when $(1)$ ground removal and clustering do not work decently, resulting in over/under segmentation; $(2)$ there are multiple similar objects nearby in a scene; $(3)$ an object is no longer in the perception range, particularly for fast-moving objects; $(4)$ point density decreases as ICP struggles to match sparse clusters.
Moreover, the naive matching strategy for cluster association (\sect{cluster_association}) does not take into consideration the one-to-one correspondence, such that a cluster might be matched multiple times.
Further, the rigid body assumption may not always hold for deformable objects, such as bendy or articulated buses and trucks.

We include typical failure cases in the supplementary material for qualitative comparison.

\section{Conclusion}
\label{sec:conclusion}

We propose a learning-free framework for scene flow estimation, particularly for LiDAR-based perception in autonomous driving. Our design is inspired by motion rigidity that assumes objects in a scene move without any deformation. To trace motion, we adopt classic ICP matching which finds the optimal transformation that aligns two clusters. We then recover the scene flow per cluster from the transformation matrix. To aid ICP, we develop a histogram-based voting for translation initialization, enabling better ICP matching. Further, we train a feedforward neural network that is capable of real-time inference using the pseudo labels from our model. We show quantitatively on Waymo, Argoverse-v2 and nuScenes the advantage of our model over other unsupervised baselines, not only from successive time steps but also from a longer time duration. Future work will fit our design into a neural network for exploiting both geometric and semantic features.


{
    \small
    \bibliographystyle{ieeenat_fullname}
    \bibliography{main}
}

\clearpage
\setcounter{page}{1}
\maketitlesupplementary

\section{ICP-Flow: cluster pairing}
This section details the optimized cluster pairing procedure introduced in \sect{pairing}, where the goal is to coarsely pair clusters that are likely to be correspondences. Further, we improve over \sect{pairing} by leveraging the cluster indices from HDBSCAN, and explain its reasoning in detail.
We start with clusters that share the same cluster index, \textit{i.e}, $\textbf{C}_{m}^{t}$ and $\textbf{C}_{n}^{t+\Delta t}$ where $ m = n$, as they are highly likely to be static or slow-moving. This is because HDBSCAN tends to group close-by points as one. 
We pair these clusters and send them to ICP matching (\sect{icp_matching}), a procedure that measures to what extent a cluster aligns with the paired one. Afterward, we reject unreliable pairs if the inlier ratio $r$ or distance $d$ exceeds the predefined threshold, \textit{i.e.}  $r < \tau_{r}$ or $d > \tau_{d}$ where $\tau_{r}$ or $\tau_{d}$ are manually defined in \sect{cluster_association}. 
We remove successfully matched pairs from the original set of clusters obtained from HDBSCAN. This way we substantially reduce the search space. 

We then process the remaining unmatched clusters after the aforementioned procedure.
We search for possible matches in a local neighborhood around $\textbf{C}_{m}^{t}$, \textit{i.e.} a square region of size $\tau_{x} \times \tau_{y}$ where $\tau_{x}$ and $\tau_{y}$ (in \textit{meters}) are the maximal translation possible within $\Delta t$ along the $x$ and $y$ dimensions. 
We pair each cluster $\textbf{C}_{m}^{t}$ with remaining clusters at time $t + \Delta t$ that lie in the predefined region. Subsequently, we feed these pairs to ICP matching (\sect{icp_matching}) and cluster association (\sect{cluster_association}) for further validation.

\section{ICP-Flow: tracking over multiple scans}
\begin{table*}[ht]
    \centering

    \resizebox{1.0\linewidth}{!}{
    \begin{tabular}{l c ccc ccc ccc}
        \toprule
        Metrics  & Label  & \multicolumn{3}{c}{Dynamic Foreground} & \multicolumn{3}{c}{Static Foreground} & \multicolumn{3}{c}{Static Background} \\
         \cmidrule(lr){3-5} \cmidrule(lr){6-8} \cmidrule(lr){9-11}
          &  & EPE ($m$) $\downarrow$ & Acc-S ($\%$) $\uparrow$ & Acc-R ($\%$) $\uparrow$ & EPE ($m$) $\downarrow$ & Acc-S ($\%$) $\uparrow$ & Acc-R ($\%$) $\uparrow$ & EPE ($m$)$\downarrow$ & Acc-S ($\%$) $\uparrow$ & Acc-R ($\%$) $\uparrow$   \\ \cmidrule(lr){1-11}
          PCA~\cite{huang2022dynamic} & \checkmark & \textbf{0.1970} & 53.31 & 77.49  & \textbf{0.0216} & \textbf{97.16} & \textbf{99.44} & \textbf{0.0289} & \textbf{97.16} & \textbf{99.44} \\
        Ours  & - & 0.2209 & \textbf{67.59} & \textbf{84.66}   & 0.0272 & 96.08 & 99.16 & 0.0711 & 96.49 & 97.96\\ 

        \cmidrule(lr){1-11}
        PCA+Kalman Tracker~\cite{huang2022dynamic} & \checkmark & 0.5860 & 36.30 & 61.60  & 0.0270 & - & - & 0.0300 & - & - \\
        Ours+Tracker   & -   &  0.1799 & 58.98 & 80.98  & 0.0341 & 88.53 & 97.73 & 0.0722 & 93.74 & 97.46   \\ 

        \bottomrule

    \end{tabular}}
    \caption{
    \textbf{Scene flow on Waymo dataset~\cite{waymo}, over a longer temporal horizon (5 consecutive frames, 0.4 seconds).} 
    Given a clip of 5 consecutive scans, we compute the flow between the first frame and the other frames. The result is averaged over all points. 
    We split models that use intermediate scans (with ``Tracker'' in their names) from others. 
    We highlight that Ours+Tracker is able to further improve the quality of scene flow by leveraging intermediate frames.  }
    \label{tab:waymo5_trakcer}
\end{table*}
We detail the design of the proposed Ours+Tracker in \sect{exp_tracker}, which estimates scene flow from a sequence of scans.
Simply speaking, we first estimate scene flow from every pair of nearby scans, thus obtaining a set of matched clusters, together with their cluster indices and transformations. Then, given a random cluster as a query, we iteratively search for its correspondence over each pair of nearby scans, starting from the current scan and stopping at the initial scan. 
Finally, we transform the query cluster sequentially by estimated transformation at each time step and recover the scene flow for a longer time duration. By this means we avoid missing matches over time. It is worth mentioning that Ours+Tracker does use intermediate frames while other models do \textit{not} use intermediate scans in \tab{waymo5}.
Additionally, we show a comparison, in \tab{waymo5_trakcer}, with PCA+Tracker\cite{huang2022dynamic}, where the learned spatio-temporal associator in the original design is replaced by a constant-velocity Kalman tracker~\cite{Weng2020_AB3DMOT, Weng2020_AB3DMOT_eccvw}. Simply speaking, the Kalman tracker solves association \textit{over time} by greedily matching the centroids of clusters based on $\textbf{L}^{2}$ distance. We directly use the result from \cite{huang2022dynamic}. The comparison between PCA and PCA+Tracker shows that the simple Kalman tracker underperforms considerably as it suffers from incorrect centroid estimation. In comparison, Our+Tracker is able to outperform PCA on dynamic foreground thanks to the ICP-based tracking.

\section{Comparison with RigidFlow~\cite{li2022rigidflow}}
We additionally compare with RigidFlow~\cite{li2022rigidflow} on the KITTI$_{o}$ dataset~\cite{liu2019flownet3d}, as both models follow the ``clustering + ICP" pipeline for flow estimation. A key difference is that RigidFlow uses a deep network for initial pose estimation, while ours uses histogram-based initialization without relying on learning from data. We report the result using the official checkpoint from authors on KITTI$_{r}$~\cite{li2021self} and using trained checkpoint by ourselves on Waymo~\cite{waymo}. Since RigidFlow does not support full point cloud inference on our device due to the high demand for GPU memory, we randomly sample a maximum of 40,000 points from each scan for inference. As shown in \tab{rigidflow_sup}, our model outperforms RigidFlow~\cite{li2022rigidflow} substantially, despite its simplicity. We did not include results on longer sequences as RigidFlow fails to produce a visually reasonable prediction.
\begin{table}[t]
    \centering
    \resizebox{1.0\linewidth}{!}{
    \begin{tabular}{l ccc ccc}
        \toprule
         Datasets & \multicolumn{3}{c} {KITTI$_{o}$} & \multicolumn{3}{c} {Waymo}  \\
          \cmidrule(lr){2-4}\cmidrule(lr){5-7}
            Metrics & EPE &  Acc-S & Acc-R & EPE (\small{Dynamic}) & EPE (\small{Static Foreground}) & EPE (\small{Static Background})\\
        \cmidrule(lr){1-7}
        RigidFlow (KITTI$_{r}$)    & 0.1192  & 40.99 & 69.77  & 0.2575 & 0.1299 & 0.2517\\    
                RigidFlow (Waymo)    & 0.2748 & 7.47 & 26.25 &0.2904 & 0.1673 & 0.3100\ \\ 
        Ours  & \textbf{0.0423} &\textbf{94.30} & \textbf{94.42} & \textbf{0.0799} & \textbf{0.0165} & \textbf{0.0270} \\ 
        \bottomrule

    \end{tabular}
    }
    \caption{
    \textbf{Comparison with RigidFlow on KITTI$_{o}$ and Waymo (0.1 seconds).} We indicate the training dataset in the bracket. Despite being simple, our model outperforms RigidFlow by a large margin, without relying on large quantities of data for training and powerful compute.
    }
    \label{tab:rigidflow_sup}
\end{table}

\section{Ablation study}
\begin{figure}[t]
    \centering
    \includegraphics[width=0.48\textwidth]{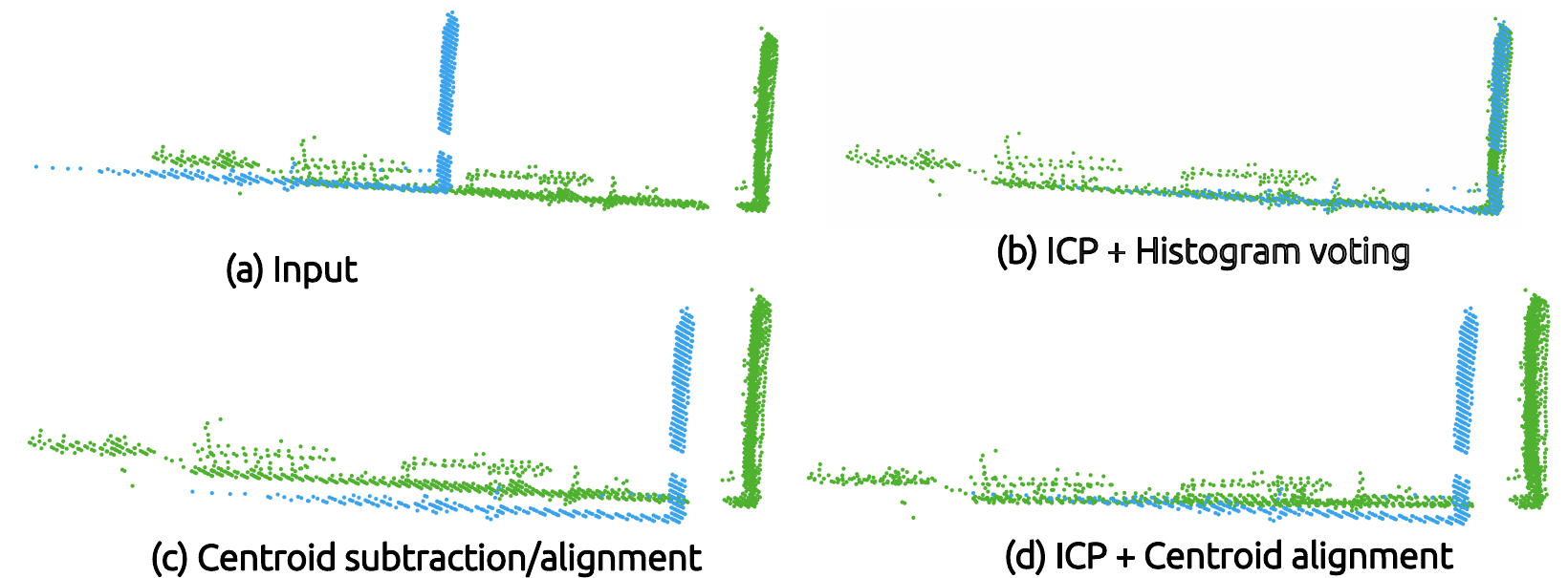}
    \caption{\textbf{ICP with centroid alignment.}  
    We show a pair of associated clusters in (a), colored in green and blue respectively. They are the bird-eye view of a moving truck. ICP fails (d) when simply subtracting the centroids (c). 
    }
    \label{fig:icp_init}
\end{figure}

We test the added value of the histogram-based initialization for ICP matching (\sect{icp_matching}) in \tab{ablation}. We compare against the commonly used centroid alignment. As shown in the result, a good initialization is essential for ICP matching as Ours \textit{(centroids)} underperforms significantly. \fig{icp_init} shows a failure case of centroid subtraction, which happens frequently over a longer temporal horizon.
Additionally, we also test the performance of our design (Ours+KISS-ICP) in the case where ego-motion information is unavailable. We use KISS-ICP~\cite{vizzo2023kiss} to estimate a relative transformation between scans. Results show a considerable performance drop on static background. Our observation aligns with~\cite{chodosh2023re} on the importance of ego motion compensation. However, it is a valid and common assumption for autonomous driving to have ego motion available. Additionally, instead of using $\argmin$ for cluster association, we also test Hungarian matching~\cite{crouse2016implementing} which yields marginally better results than the default setup.

\begin{table}[t]
    \centering

    \resizebox{1.0\linewidth}{!}{
    \begin{tabular}{l ccc}
        \toprule
          & Dynamic & Static & Static \\
            & Foreground &  Foreground & Background \\
        \cmidrule(lr){1-4}
        
        Ours  & 0.2209 & 0.0272 & 0.0711 \\ 
        Ours \textit{(centroid alignment)}      & 0.3511 & 0.0789 & 0.1861\\ 
        Ours \textit{(Hungarian Matching)}      & \textbf{0.2163}  & \textbf{0.0260} & \textbf{0.0681}\\ 
        Ours+KISS-ICP~\cite{vizzo2023kiss}     & 0.2617  & 0.0572  &  0.3386\\
        \bottomrule

    \end{tabular}}
    \caption{
    \textbf{Ablation study.} We report EPE errors on Waymo over 5 consecutive frames~\cite{waymo, huang2022dynamic}. Without the histogram-based initialization, the performance decreases substantially. 
    Precise ego-motion is also critical for scene flow, particularly for static background.
    When replacing $\argmin$ by Hungarian matching~\cite{crouse2016implementing} during cluster assignment, our model yields marginally better results. 
    }
    \label{tab:ablation}
\end{table}

\section{Visualization}
\begin{figure*}[!ht]
    \centering
    \includegraphics[width=0.7\textwidth]{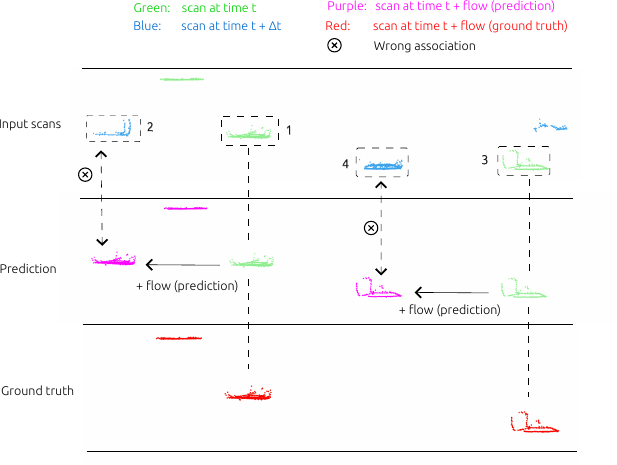}
    \caption{\textbf{Visualization of predicted scene flow.}  
    We qualitatively compare our prediction to the ground truth. For better visualization, we crop the region of interest from the entire scan.
    We plot the input scans at time $t$ and $t+\Delta t$, namely $\textbf{X}_{t}$ and $\textbf{X}_{t+\Delta t}$, in green and blue, respectively. We color the flow-compensated scan at time $t$, namely $\textbf{X}_{t}^{\prime}$, in purple by adding the predicted scene flow $\textbf{F}_{t}$ to $\textbf{X}_{t}$. In comparison, we use red to indicate the flow-compensated scan at time $t$, namely $\textbf{X}_{t}^{*}$, by adding the \textit{ground truth} flow. 
    The left figure is composed of $\textbf{X}_{t}$, $\textbf{X}_{t+\Delta t}$ and $\textbf{X}_{t}^{\prime}$. ICP-Flow is able to output reasonable predictions once the blue and purple points align (\textit{i.e.} overlap) with each other. However, ICP-Flow fails in certain scenarios by associating the wrong clusters, as indicated by the box on the top. We highlight this failure in the right figure, where \xmark ~denotes a wrong association. As indicated by the dashed lines on the left, ICP-Flow associates clusters $1$ and $2$ (or $\textbf{C}_{1}^{t}$ and $\textbf{C}_{2}^{t+\Delta t}$ ), and estimates a transformation that best aligns them. Unfortunately, $\textbf{C}_{1}^{t}$ remains static within $\Delta t$ according to the ground truth (in red). Similarly, we observe that $\textbf{C}_{3}^{t}$ and $\textbf{C}_{4}^{t+\Delta t}$ are also falsely associated. Interestingly, after careful examination, we find this an annotation error in the preprocessed Waymo dataset~\cite{huang2022dynamic}, as explained in \fig{failure_353_raw}.
    }
    \label{fig:failure_353}
\end{figure*}

\begin{figure*}
    \centering
    \begin{tabular}{c}
    \includegraphics[width=0.8\textwidth]{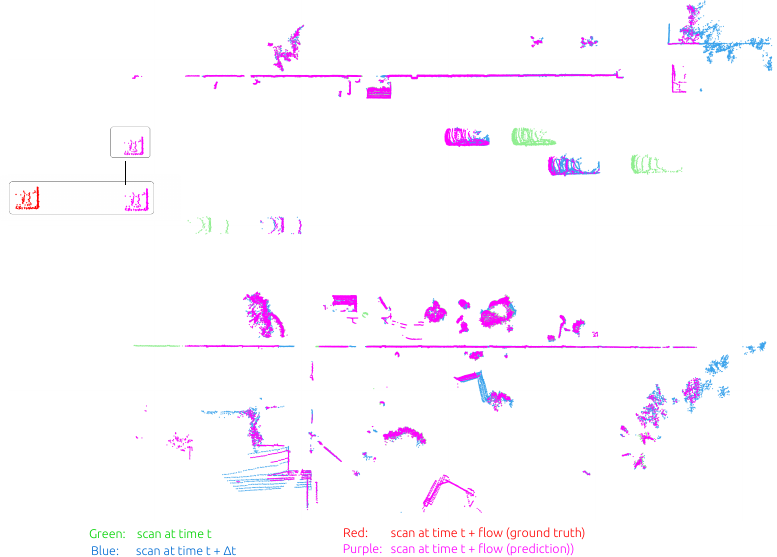}  
    \end{tabular}
    \caption{\textbf{Failure case.} We show another failure case where a cluster moves out of the perception range, as indicated in the box. Thus ICP-Flow fails to associate and outputs zero scene flow, \textit{i.e.} the cluster moves identically to the ego autonomous vehicle. This often leads to substantially large errors for dynamic foreground.}
    \label{fig:failure_196}
\end{figure*}

\begin{figure*}
    \centering
    \begin{tabular}{c}
    \includegraphics[width=0.8\textwidth]{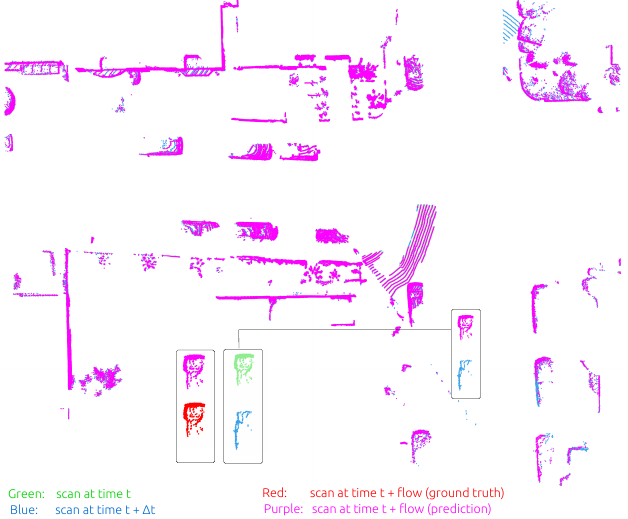}  
    \end{tabular}
    \caption{\textbf{Failure case.} We show a failure case where occlusion happens. We highlight this failure in boxes, where our model predicts zeros for the given cluster (in green), as the purple and green points overlap seamlessly. This results from $(1)$ low inlier ratio, as the blue cluster at time $t+\Delta t$ consists of much fewer points than the green cluster at time $t$; (2) partial occlusion, as we are unable to observe the blue cluster from a similar view, thus making it hard to match with the green cluster. }
    \label{fig:failure_169}
\end{figure*}

We visualize the predicted scene flow from our model and highlight several failure cases in \fig{failure_353}, \fig{failure_169}, and \fig{failure_196}. These qualitative results show the capability of ICP-Flow to extract scene flow in various scenarios reliably.

\begin{figure*}
    \centering
    \begin{tabular}{c}
    \includegraphics[width=0.55\textwidth]{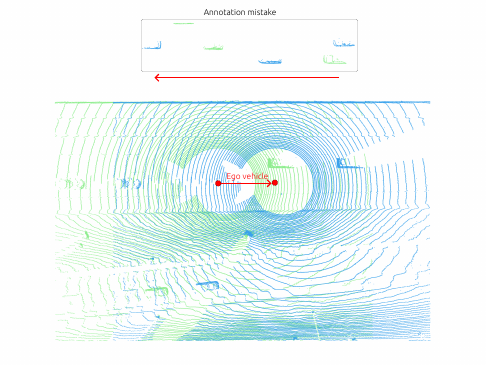}  
    \end{tabular}
    \caption{\textbf{Visualization of the original scans from \fig{failure_353}, after ego-motion compensation.} After careful examination, we find that \fig{failure_353} is not perfectly annotated and ICP-Flow is actually making a reasonable prediction. We highlight the clusters that a visual examiner intends to associate in boxes, based on the observation that they are heading from right to left (indicated by the red arrow below the box). However, in the preprocessed Waymo dataset~\cite{huang2022dynamic}, these points (in green and inside the box) are labeled as static (\textit{i.e., } without having correspondences), which we assume to be a mistake during preprocessing. We manually examined numerous examples and did not find other annotation errors.}
    \label{fig:failure_353_raw}
\end{figure*}

\end{document}